# Toward Optimal Sampling Rate Selection and Unbiased Classification for Precise Animal Activity Recognition

Axiu Mao [a], Meilu Zhu [b,c*], Lei Shen [a], Xiaoshuai Wang [d], Tomas Norton [e], Kai Liu [f*]

[a] School of Communication Engineering, Hangzhou Dianzi University, Hangzhou 310018, China

[b] Department of Mechanical Engineering, City University of Hong Kong, Hong Kong SAR 999077, China

[c] Department of Electrical and Computer Engineering, The University of Hong Kong, Hong Kong SAR 999077, China

[d] College of Biosystems Engineering and Food Science, Zhejiang University, Hangzhou 310058, China

[e] Department of Biosystems, Division Animal and Human Health Engineering, M3-BIORES, Katholieke University of Leuven, Kasteelpark Arenberg 30, 3001 Heverlee, Belgium

[f] Department of Infectious Diseases and Public Health, Jockey Club College of Veterinary Medicine and Life Sciences, City University of Hong Kong, Hong Kong SAR 999077, China

* Corresponding author. Email address: Meilu Zhu: meiluzhu2-c@my.cityu.edu.hk; +852 64836454.

Kai Liu: kailiu@cityu.edu.hk; +852 3442-5295.

## Abstract

With the rapid advancements in deep learning techniques, wearable sensor-aided animal activity recognition (AAR) has demonstrated promising performance, thereby improving livestock management efficiency as well as animal health and welfare monitoring. However, existing research often prioritizes overall performance, overlooking the fact that classification accuracies for specific animal behavioral categories may remain unsatisfactory. This issue typically stems from suboptimal sampling rates or class imbalance problems. To address these challenges and achieve high classification accuracy across all individual behaviors in farm animals, we propose a novel Individual-Behavior-Aware Network (IBA-Net). This network enhances the recognition of each specific behavior by simultaneously customizing features and calibrating the classifier. Specifically, considering that different behaviors require varying sampling rates to achieve optimal performance, we design a Mixture-of-Experts (MoE)-based Feature Customization (MFC) module. This module adaptively fuses data from multiple sampling rates, capturing customized features tailored to various animal behaviors. Additionally, to mitigate classifier bias toward majority classes caused by class imbalance, we develop a Neural Collapse-driven Classifier Calibration (NC$^3$) module. This module introduces a fixed equiangular tight frame (ETF) classifier during the classification stage, maximizing the angles between pair-wise classifier vectors and thereby improving the classification performance for minority classes. To validate the effectiveness of IBA-Net, we conducted experiments on three public datasets covering goat, cattle, and horse activity recognition. The results demonstrate that our method consistently outperforms existing approaches across all datasets, achieving the highest values of 93.17%, 87.17%, and 91.71% in accuracy, F1-score, and recall for the goat dataset, 91.42%, 90.62%, 88.06%, and 93.64% in accuracy, F1-score, precision, and recall for the cattle dataset, and 92.22%, 84.34%, 84.05%, and 84.65% in accuracy, F1-score, precision, and recall for the horse dataset. Notably, the recall values exhibit substantial enhancements, surpassing the baseline method by 16.98%, 15.78%, and 4.91% on the goat, cattle, and horse datasets, respectively. These results confirm that IBA-Net effectively enhances classification accuracy across all animal behavioral categories, particularly for minority classes, while aligning with the initial goal of achieving high recall values and maximizing overall classification performance.

## 1. Introduction

Recent advancements in deep learning are accelerating the development of automated and precise animal activity recognition (AAR) via wearable sensors. This enables real-time behavioral monitoring and early disease detection, thereby enhancing livestock management and welfare (Arablouei et al., 2024; Mao et al., 2023a). Literature has demonstrated the promising performance of deep learning models in distinguishing various animal activities. Arablouei et al. (2023, 2021) and Hosseininoorbin et al. (2021) proposed multilayer perceptron (MLP) algorithms using sensor data, achieving accuracies of around or above 90% in classifying cattle behaviors like grazing, ruminating, and resting. Convolutional neural networks (CNNs) are currently the most widely applied models in AAR tasks, arriving at accuracies over 90% in identifying behaviors across various species, such as the standing and walking of horses (Eerdekens et al., 2021; Mao et al., 2022b), the ruminating and salt-licking of cattle (Bloch et al., 2023; Li et al., 2021), the grazing and activity levels of sheep (Kleanthous et al., 2022), and the eating and nursing of pig (Pan et al., 2023). Recurrent neural networks (RNNs), advantageous for modeling sequential data like sensor inputs, are also increasingly applied, sometimes in combination with CNNs, often outperforming pure CNN- or RNN-based models (Kim and Moon, 2022; Liseune et al., 2021).

Despite the satisfactory overall performance of deep learning in the AAR task, some specific behavioral categories may exhibit undesirable classification accuracies due to confusion among distinct classes. This phenomenon normally arises from either suboptimal sampling rates or class imbalance problems. Firstly, the sampling rate of data is a crucial factor impacting the performance of animal behavior classification, as it directly influences the discriminative degree of the extracted features. Although existing studies have investigated the optimal sampling rate for classification, they all select it based on overall performance without considering that this sampling rate may be sub-optimal for specific behavioral categories. For instance, Walton et al. (2018) assessed sampling rates (8 Hz, 16 Hz, and 32 Hz) for recognizing sheep activity based on tri-axial acceleration and angular velocity data, finding that the highest overall accuracy of 95% was achieved at 32 Hz. Eerdekens et al. (2021, 2020) and Mao et al. (2023b) evaluated the impact of various sampling rates, specifically 100 Hz, 50 Hz, 25 Hz, and 12.5 Hz, on the horse behavior classification using tri-axial acceleration and angular velocity data. Their results indicated that a sampling rate of 25 Hz yielded the best overall performance. These methods that select sampling rates by maximizing overall classification accuracy often fail to guarantee ideal performance for each behavior. This is mainly due to varying movement patterns (e.g., periodicity) among behaviors, requiring different sampling rates for optimal performance. Therefore, investigating how the sampling rate impacts the classification accuracy of each behavior, as well as determining the optimal one for each category within an AAR model, is an area of research worth pursuing.

Secondly, the class imbalance problem often stems from the inconsistent frequencies and durations of various behaviors, due to the animal's inherent physiologies. It has been discovered that classifiers trained on imbalanced datasets are often biased toward majority classes, leading to high misclassification rates for rare categories. For example, Arablouei et al. (2023) utilized an MLP-based network to recognize five daily cattle behaviors, achieving an overall accuracy of 97.75%. Nevertheless, lower classification accuracies were obtained for specific behaviors, with walking at 67.69% and drinking at 48.08%, which is directly related to the fact that the sample sizes for these two behaviors only accounted for 7.61% and 4.97% of the total, respectively (Fig. 1a). This aligns with our previous research on horse activity recognition, which reported an overall accuracy of 93.37%, but a significantly lower accuracy of 24.75% for the "walking-natural" behavior, as depicted in Fig. 1b (Mao et al., 2021). Current research has attempted to mitigate the class imbalance problem primarily through resampling to balance class distribution (e.g., over-sampling and under-sampling (Zhang et al., 2021), generative adversarial networks (Suh et al., 2021)) or reweighting the loss function to increase the penalty for the minority class (e.g., cost-sensitive loss (Khan et al., 2018), class-balanced focal loss (Cui et al., 2019), adaptive class suppression loss (Wang et al., 2021)). It's obvious that these approaches exclusively focus on enhancing classifier training by optimizing external factors while ignoring the desired inner structure of a well-trained classifier in balanced conditions. Thus, exploring the internal structure of a classifier under imbalanced and balanced conditions is necessary before seeking solutions.

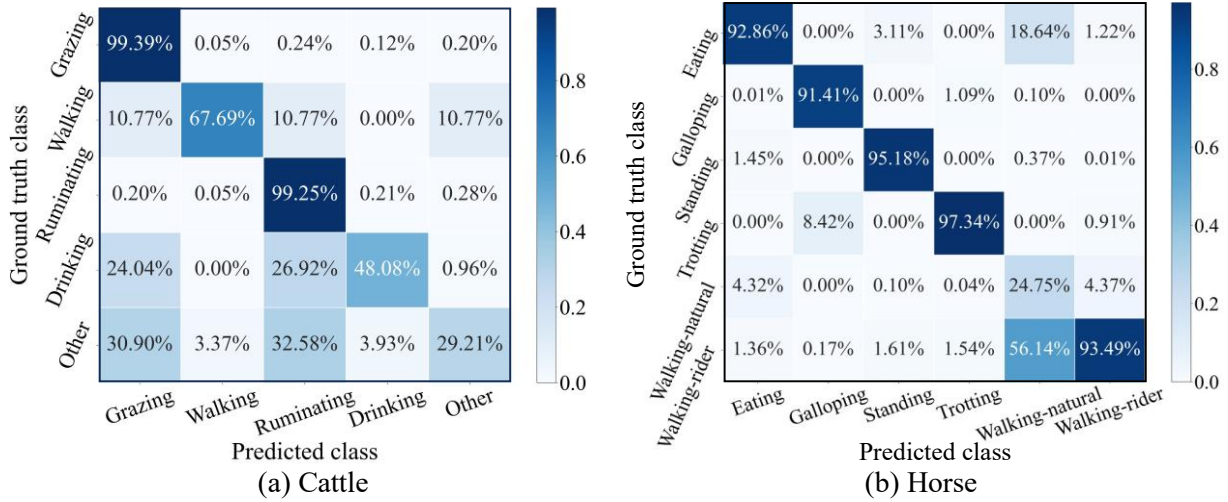

**Fig. 1**. Confusion matrices of cattle activity recognition (a; Arablouei et al. (2023)) and horse activity recognition (b; Mao et al. (2021)).

In this study, we propose a novel method, namely the Individual-Behavior-Aware Network (IBA-Net), primarily focusing on classification across all individual behavioral categories in farm animals. Considering that the sampling rates required to achieve optimal performance vary among different animal behaviors, we design a Mixture-of-Experts (MoE)-based Feature Customization (MFC) module that can adaptively fuse data from multiple sampling rates, offering the potential to capture personalized and discriminative features for each behavior. To mitigate the classifier's bias in classifying behaviors due to class imbalance, we develop a Neural Collapse-driven Classifier Calibration (NC$^3$) module, which introduces a fixed equiangular tight frame (ETF) classifier in the classification stage to maximize the angles between pair-wise classifier vectors. To validate the classification capability of IBA-Net across all behaviors, we compare it against several existing methods on three public datasets: a goat dataset (Kamminga et al., 2018), a cattle dataset (Li et al., 2021), and a horse dataset (Kamminga et al., 2019). To summarize, our contributions are as follows.

- This work introduces a paradigm shift from optimizing overall classification accuracy to ensuring high-fidelity recognition for each individual behavior in farm animals. Our proposed IBA-Net is the first to achieve this by synergistically customizing behavior-aware features and calibrating the classifier, establishing a new direction for wearable sensor-aided AAR.

- We design an MFC module that adaptively fuses data from multiple sampling rates. This enables the automatic extraction of optimal, behavior-specific patterns without relying on pre-known labels, effectively addressing the challenge of divergent optimal sampling rates across different activities.

- Motivated by the desired inner structure of a well-trained classifier in balanced conditions, we devise an NC$^3$ module that incorporates a fixed, structurally optimal ETF classifier. This innovation directly enforces maximal separation between class prototypes, effectively mitigating classifier bias toward majority classes and significantly improving recognition of rare behaviors.

- Experiments conducted on three public farm animal datasets demonstrate that IBA-Net achieves state-of-the-art performance. It shows substantial and consistent improvements, particularly in recall for minority classes, confirming its practical efficacy and robustness.

## 2. Methods

### 2.1 Motivation and problem formulation

#### 2.1.1 How does the sampling rate affect performance across various behaviors?

To investigate how the sampling rate affects performance across various behaviors, we visualize the recall confusion matrices of the goat behavioral classification model at various sampling rates (50 Hz, 25 Hz, and 12.5 Hz) in Figure 2. The recall metric represents the percentage of samples that are correctly classified (Mao et al., 2022a). It is apparent that the optimal sampling rate varies depending on the specific behavior being analyzed. Specifically, the "running" behavior attains the highest accuracy at a sampling rate of 50 Hz, with accuracy decreasing as the sampling rate is lowered. Conversely, the "standing" and "trotting" behaviors perform better at a relatively lower sampling rate, and their accuracies decline as the sampling rate increases. In addition, the "grazing" and "walking" behaviors exhibit superior performance at a sampling rate of 25 Hz compared to the other two considered sampling rates.

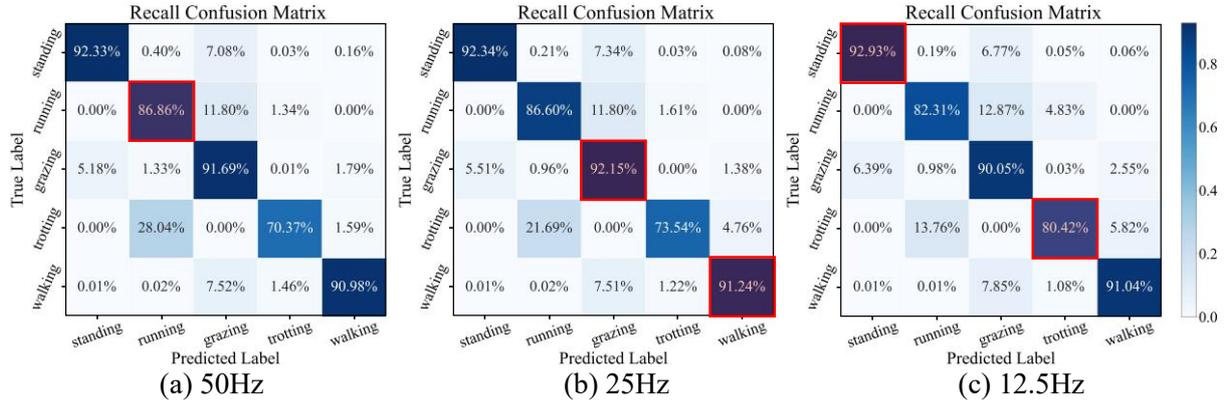

(a) 50Hz      (b) 25Hz      (c) 12.5Hz

**Fig. 2**. Confusion matrices of the goat behavioral classification model at various sampling rates, including 50 Hz (a), 25 Hz (b), and 12.5 Hz (c).

Based on the above results, we analyze that the performance of "running" is inferior at lower sampling rates. This is primarily attributed to the fact that lower rates can lead to the loss of critical details, such as peaks and troughs, and alterations in the intrinsic regularity of signals, e.g., their periodicity. Consequently, the distinction between "running" and other behaviors, such as "trotting", becomes blurred, increasing the likelihood of misclassification, as demonstrated in our earlier research (Mao et al., 2023b). Conversely, the accuracy of "trotting" and "standing" decreases at higher sampling frequencies. This may be because their behavioral patterns are already adequately represented when sampled at 12.5Hz. Increasing the sampling rate captures more noise, which could potentially lead the model to misidentify the patterns as those associated with "running" and "grazing".

An intuitive idea would be to integrate data from multiple sampling rates to ensure effective and tailored information for different behaviors. With this objective in mind, our research will delve into the methodologies for achieving this integration.

### 2.1.2 What occurs within the internal structure of a classifier under imbalanced and balanced conditions?

To understand why the class imbalance problem leads to higher misclassification rates for minority classes, we endeavor to explore the internal structure of the trained classifier by visualizing the pair-wise angles between the classifier vectors across three independent training folds in Fig. 3. Generally, smaller angles indicate greater proximity between classifier weight vectors in the embedding space. We can observe that the angles among trotting, running, and walking are relatively small, all measuring below 90°. Meanwhile, we notice that the dataset exhibits a clear class imbalance, with the minority classes "trotting", "running", and "walking" constituting only 0.44%, 0.87%, and 20.19% of the total dataset, respectively—proportions significantly lower than those of the other two behavioral classes. Such a phenomenon aligns with the assertion made by Fang et al. (2021) that classifier vectors of minority classes would merge in cases of class imbalance.

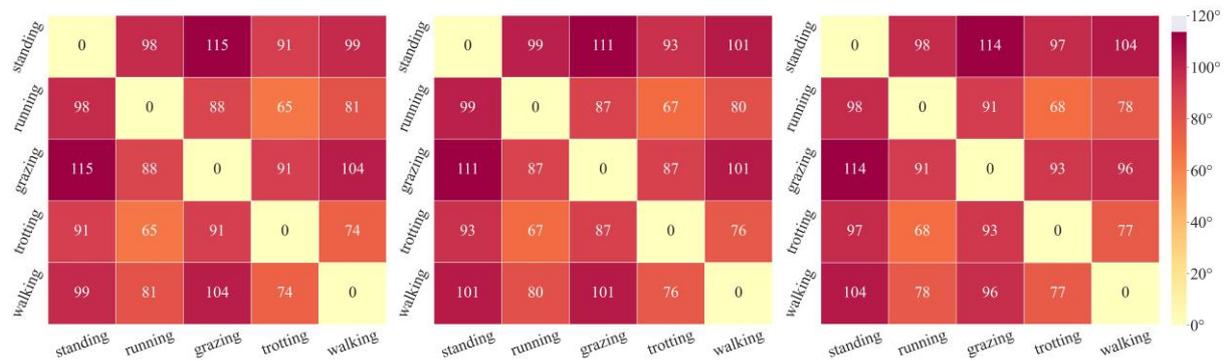

**Fig. 3**. Illustration of the pair-wise angles between the classifier vectors across three training folds.

Regarding the internal structure of the aforementioned classifiers under class imbalance, existing research has also explored their internal structure when balance is achieved. Neural collapse (Papyan et al., 2020), an emerging discovery, has shed light on the properties of a well-trained classifier under class balance conditions. It reveals a highly symmetric phenomenon that, at the terminal phase of training on a balanced dataset, the last-layer features collapse to class-wise means (feature prototypes), and both feature prototypes and classifier vectors converge to an optimal simplex equiangular tight frame (ETF). A simplex ETF represents a geometric structure that maximizes the pair-wise angles among $K$ vectors in $d$ dimensions, where $d \geq K - 1$, and all vectors possess an equal $\ell_2$ norm. Here, $K$ denotes the number of behavioral categories. Figure 4 illustrates the simplex ETF in scenarios of class balance, with $K = 4, d = 3$.

Motivated by these observed empirical results and findings, we aim to introduce an ETF classifier as a regularization term into the original classification stage, with the goal of maximizing the angles between pair-wise classifier vectors as much as possible.

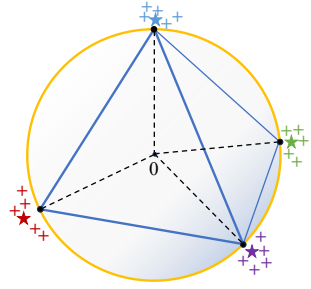

**Fig. 4**. The three-dimensional feature spaces and class centers under scenarios of class balance.

## 2.2 The proposed method

### 2.2.1 Overview of the Individual-Behavior-Aware Network

The overall workflow of the proposed Individual-Behavior-Aware Network (IBA-Net) is illustrated in Fig. 5. Considering that the optimal sampling rate varies across different behaviors, the IBA-Net integrates the data with multiple distinct sampling rates and attempts to derive customized and effective features for each behavior by a newly devised MoE-based Feature Customization (MFC) module. Subsequently, the extracted features are fed into a novel Neural Collapse-driven Classifier Calibration (NC$^3$) module, which aims to achieve fair and accurate behavior classification under class imbalance scenarios. The details of the newly designed MFC and NC$^3$ modules are described below.

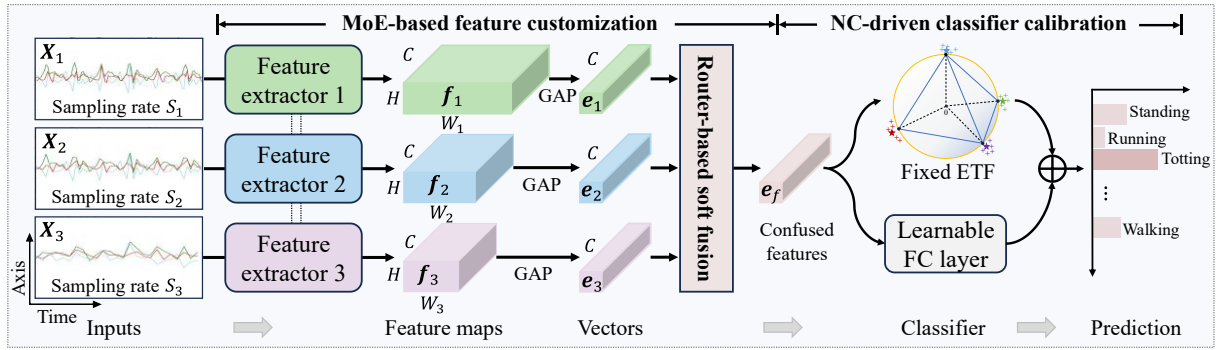

**Fig. 5**. Overview of the proposed Individual-Behavior-Aware Network (IBA-Net) for animal activity recognition. Herein, "MoE", "NC", "ETF", and "FC" denote mixture-of-experts, neural collapse, equiangular tight frame, and fully connected, respectively.

## 2.2.2 MoE-based feature customization

Selecting an optimal sampling rate is crucial for the performance of wearable sensor-aided AAR. However, solely applying a unified sampling rate is challenging to ensure ideal performance for every individual behavior, as the sampling rates required to achieve optimal performance vary among different behaviors. An intuitive idea is to integrate data from multiple sampling rates, which could provide opportunities to derive personalized movement patterns for various behaviors. One straightforward way is to select data at the optimal sampling rate corresponding to a specific behavior, but this method heavily relies on pre-known behavior labels, which are only feasible during the training phase and not applicable during the testing phase. Therefore, we design an MFC module that adaptively fuses data from different sampling rates, without the need for pre-known behavior labels, to provide customized features tailored to various behaviors.

As illustrated in Fig. 5, we take the inclusion of three sampling rates, i.e., $S_1, S_2, S_3$, as an illustrative example, and assume that the values of these three sampling rates adhere to the following order: $S_1 > S_2 > S_3$. Specifically, data obtained from these sampling rates, i.e., $X_1, X_2$, and $X_3$, are initially fed into three respective feature extractors to capture unique features, where all feature extractors share the same network architecture. These extracted features are represented as $f_1 \in R^{C \times H \times W_1}, f_2 \in R^{C \times H \times W_2}$, and $f_3 \in R^{C \times H \times W_3}$, where $C$ denotes the channel number, $H$ corresponds to the axis dimension, $W_1, W_2, W_3$ represent the temporal dimensions ($W_1 > W_2 > W_3$), and collectively $H \times W_1, H \times W_2, H \times W_3$ refer to the spatial dimension of the feature maps. Due to the inconsistency in the aforementioned temporal dimensions, we apply the global average pooling operation to each feature map along their respective spatial dimensions ($H \times W_1, H \times W_2, H \times W_3$), producing three feature vectors, i.e., $e_1 \in R^C, e_2 \in R^C$, and $e_3 \in R^C$. This approach resolves the dimensional conflicts introduced by different sampling rates, ensuring seamless processing in the subsequent MFC module. This means that the differences in sampling rates do not affect the processing within the MFC module, as the conflict induced by the varying rates has been resolved. Subsequently, the three features are fed into a router-based soft fusion layer for adaptive fusion (Fig. 6), yielding a fused feature ( i.e., $e_f \in R^C$).

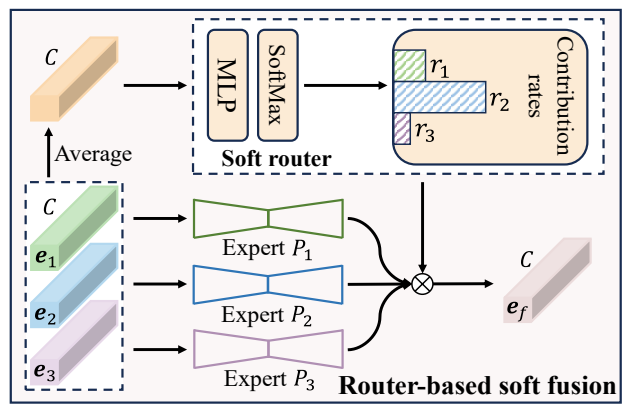

**Fig. 6**. The architecture of the MoE-based Feature Customization (MFC) module.

As shown in Fig. 6, the router-based soft fusion layer incorporates a set of expert networks alongside a routing network and can adaptively assess the contribution degrees of features extracted at various sampling rates and perform soft fusion on these features. Since the three feature vectors, denoted as $e_1 \in R^C$, $e_2 \in R^C$, and $e_3 \in R^C$, are extracted from data sampled at three distinct rates using three corresponding expert networks that share the same architecture, the receptive fields within these features span across different time scales. As a result, these three features contain varying levels of information granularity in the temporal dimension. Features with coarser granularity capture broader, more macroscopic behavioral patterns, whereas those with finer granularity detail more refined, intricate patterns. To better fuse these three features with various information granularity, we employ three respective projection experts, i.e., $P_1, P_2, P_3$, with identical structures to align them into a unified embedding space. Each projector expert is a two-layer hourglass-shaped MLP, where each fully connected layer is followed by a Gaussian Error Linear Unit (GELU) activation function, owing to GELU's ability to mitigate the vanishing gradients issue (Hendrycks and Gimpel, 2016). Meanwhile, a soft router $R_{soft}$ is designed to adaptively evaluate the contribution rate of each feature vector based on its respective values. Specifically, we first compute the average of all three feature vectors to obtain a single combined vector. This vector is then processed through a two-layer MLP, followed by a SoftMax operation with a temperature parameter $\tau$, resulting in three contribution rates, denoted as $\{r_1, r_2, r_3\}$:

$$o = MLP(Avg(e_1, e_2, e_3)), o \in R^3, \tag{1}$$

$$r_i = \frac{e^{o_i/\tau}}{\sum_{j \in \{1,2,3\}} e^{o_j/\tau}}, \ i = 1,2,3, \tag{2}$$

where parameter $\tau$ controls the involvement degrees of features across different sampling rates. Then, the final fused feature $e_f \in R^C$ can be formulated as the weighted sum of each projected feature, i.e.,

$$e_f = \sum_{i \in \{1,2,3\}} r_i * P_i(e_i), \tag{3}$$

where the weights are their corresponding contribution rates $\{r_1, r_2, r_3\}$. Through the operation defined in Eq. (3), the MFC module effectively ensures that the features acquired from the optimal sampling rate are maximized, while simultaneously incorporating complementary information derived from data sampled at other rates.

*2.2.3 Neural collapse-driven classifier calibration*

A classifier based solely on fully connected (FC) layers often faces significant challenges under class imbalance scenarios, particularly with minority classes. Classifier vectors of minority classes tend to merge, leading to minority collapse, which degrades test performance (Fang et al., 2021). Recent studies have shown that under class balance scenarios, neural networks aim to achieve an optimal state known as neural collapse, where both feature prototypes and classifier vectors converge toward an ETF structure characterized by maximal pair-wise angles (Zarka et al., 2021). This ETF structure is considered an ideal target for neural network training, as it maximizes the separation between classes. However, preliminary experiments indicate that directly applying the ETF structure to class imbalance scenarios is not feasible. To address this challenge, we propose an NC³ module that integrates a fixed ETF classifier with a learnable FC layer-based classifier, as shown in Fig. 5. This module establishes an ETF classification branch with maximal pair-wise angles and keeps its weight vectors fixed during training, allowing only the other classification branch to train normally. By combining the ETF structure with the FC layer, this approach maximizes the angles between pair-wise classifier vectors as much as possible, effectively mitigating the minority collapse issue caused by class imbalance. Notably, this calibration operates primarily in the geometric space of class representations, by enforcing an optimal ETF structure among classifier weights, rather than through post-hoc adjustment of the output probabilities.

Within the ETF classification branch, we randomly synthesize vectors at the beginning of the training and ensure they adhere to the simplex ETF properties. Concretely, assuming that we have a set of $M$ vectors $\boldsymbol{V} = [\boldsymbol{v}_1, \boldsymbol{v}_2, \cdots \boldsymbol{v}_M] \in R^{d \times M}$, where $d$ denotes the feature dimension and $M$ represents the class number, with the condition that $d \geq M - 1$. This vector set is deemed to be a simplex ETF if:

$$\boldsymbol{V} = \sqrt{\frac{M}{M-1}} \mathbf{U} (\mathbf{I}_M - \frac{1}{M} \mathbf{1}_M \mathbf{1}_M^T), \tag{4}$$

where $\mathbf{U} \in R^{d \times M}$ allows for a rotation and satisfies $\mathbf{U}^T \mathbf{U} = \mathbf{I}_M$, where $\mathbf{I}_M$ is the identity matrix, and $\mathbf{1}_M$ denotes an all-ones vector of length $M$. All vectors within the simplex ETF have an equal $\ell_2$ norm and the same pair-wise angle, i.e.,

$$\boldsymbol{v}_i^T \boldsymbol{v}_j = \frac{M}{M-1} \delta_{i,j} - \frac{1}{M-1}, \forall i,j \in [1,2,\ldots,M], \tag{5}$$

$$\delta_{i,j} = \begin{cases} 1, & i = j \\ 0, otherwise \end{cases}. \tag{6}$$

The pair-wise angle of $-\frac{1}{M-1}$ represents the maximal equiangular separation among $M$ vectors. Considering that the normalized class-wise feature prototypes $\{\widetilde{\boldsymbol{h}}_m = \frac{\boldsymbol{v}_m}{\|\boldsymbol{v}_m\|} \in R^d\}_{m=1}^M$ align with their respective classifier weights, we regard these normalized vectors $\{\widetilde{\boldsymbol{h}}_m \in R^d\}_{m=1}^M$ as the weights of the ETF classifier and maintain them fixed during training.

On the other hand, the fused feature $\boldsymbol{e}_f \in R^C$ obtained from Eq. (3) is projected via a projector $g$ with the intention of mapping it into the ETF feature space, yielding the feature $\tilde{\boldsymbol{e}}_f \in R^d$. Given that high-dimensional vectors are more prone to being orthogonal, which presents a challenge in collapsing them into an ETF with maximal angles (Li et al., 2023), we set $d = M$ to regulate the dimensionality, thereby aiding in neural collapse. Then, the feature $\tilde{\boldsymbol{e}}_f$ is normalized to $\bar{\tilde{\boldsymbol{e}}}_f \in [-1,1]$, which is subsequently multiplied by the predefined regarded classifier weights $\{\widetilde{\boldsymbol{h}}_m\}_{m=1}^M$ to produce the logits $\acute{\boldsymbol{Z}} = [\acute{z}_1, \acute{z}_2, \ldots, \acute{z}_M]$. The operations are shown as follows:

$$\acute{z}_m = \mu \widetilde{\boldsymbol{h}}_m^T \bar{\tilde{\boldsymbol{e}}}_f, \tag{7}$$

$$\bar{\tilde{\boldsymbol{e}}}_f = \frac{\tilde{\boldsymbol{e}}_f}{\|\tilde{\boldsymbol{e}}_f\|}, \tilde{\boldsymbol{e}}_f = g(\boldsymbol{e}_f), \tag{8}$$

where $\mu$ denotes the learnable temperature parameter used to scale the features' product $\acute{Z}$, since both $\widetilde{h}_m$ and $\bar{\bar{e}}_f$ have a limited range of [-1,1].

Within the learnable FC layer-based classifier, we initialize its parameters randomly and allow them to undergo normal training during the training process. Denoting its weights as $W = [w_1, w_2, ..., w_M]$, it processes the fused feature $e_f \in R^C$ to produce the logits $\grave{Z} = [\grave{z}_1, \grave{z}_2, ..., \grave{z}_M]$, as shown below:

$$\grave{z}_m = w_m^T e_f. \tag{9}$$

Afterward, the output logits $\acute{Z}$ within the ETF classification branch and $\grave{Z}$ within the learnable FC layer-based classifier are linearly combined to form the final outputs:

$$Z = k * \acute{Z} + (1 - k) * \grave{Z}, 0 < k \leq 1. \tag{10}$$

where $k$ is the weight coefficient controlling the extent to which the ETF classification parameters are fixed. Through introducing the fixed ETF classification parameters, the angles between pair-wise classifier vectors can be effectively increased, thus alleviating the minority collapse issue in the class imbalance scenarios.

*2.2.4 Model optimization*

This study employs class-balanced focal loss (CB_FL) as the loss function, which has been proven effective in addressing the class imbalance problem in our previous research (Mao et al., 2021). This kind of loss not only emphasizes samples from minority classes, mitigating their tendency to be overshadowed during optimization, but also focuses on samples that are hard to distinguish. Let $Z = [z_1, z_2, ..., z_M]$ denote the final output logits, the class-balanced focal loss $\mathcal{L}_{cbfocal}$ can be formulated as:

$$\mathcal{L}_{cbfocal} = -\frac{1-\beta}{1-\beta^{n_y}} \sum_{m=1}^{M} (1 - p_m^t)^\gamma log(p_m^t), \tag{11}$$

$$\text{with } p_m^t = \frac{1}{1+e^{-z_m^t}}, \quad z_m^t = \begin{cases} z_m, & \text{if } m = y. \\ -z_m, & \text{otherwise.} \end{cases} \tag{12}$$

where $n_y$ denotes the number of samples with the ground-truth label $y$, $\beta \in [0,1)$ controls the growth rate of the effective number of samples belonging to $y$ as $n_y$ increases, and $\gamma \geq 0$ smoothly adjusts the rate at which easy samples are down-weighted (Cui et al., 2019). Herein, we set $\beta$ and $\gamma$ to 0.9999 and 0.5, respectively. Notably, our work deals with a multi-class classification problem. When applying the CB_FL, we employ a "one-against-rest" strategy (Vural and Dy, 2004), which frames the problem as a series of binary classification tasks—one for each class, where the target class is distinguished against all remaining classes collectively. For every class, the probability of a sample belonging to it is computed via sigmoid activation, and the CB_FL is used to balance the influence of minority classes and hard examples across these binary tasks. The final behavioral category is determined by selecting the class with the highest resulting probability.

## 3. Datasets and design of experiments

### 3.1 Datasets

Our proposed method is tested using three open-source datasets collected from goats (Kamminga et al., 2018), cattle (Li et al., 2021), and horse (Kamminga et al., 2019), respectively.

***Goat dataset.*** The goat dataset was collected from five goats across two farms: three domestic pygmy goats from one farm and two larger, more wild goats from the other, as shown in Fig. 7. Each goat wore a collar attached to six triaxial accelerometers and triaxial gyroscopes fixed in different orientations, with a sampling rate of 100 Hz. A total of 42,943 2-second data samples were selected, resulting in a tensor of $1 \times 36 \times 200$ for each sample. Figure 8a depicts the class distribution of these samples, covering five main activities, i.e., standing, running, grazing, trotting, and walking, with percentages of 43.15%, 0.87%, 35.35%, 0.44%, and 20.19%, respectively. It's obvious that this dataset is severely imbalanced, with an imbalance ratio of 98.05.

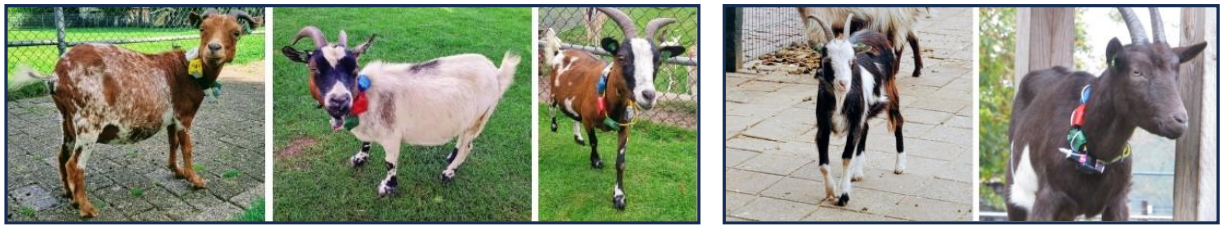

(a) Three smaller Pygmy goats    (b) Two larger goats

**Fig. 7**. Images of goats equipped with wearable sensors in two different farms (Kamminga et al., 2018).

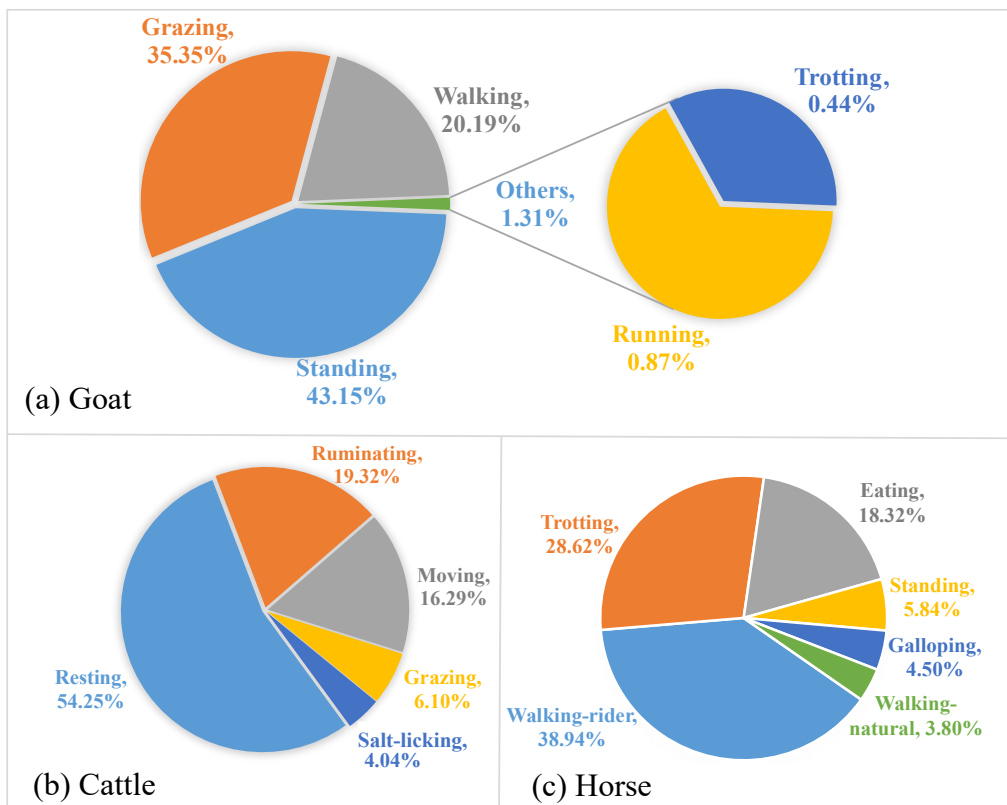

**Fig. 8**. Class distribution of the goat dataset (a), the cattle dataset (b), and the horse dataset (c).

***Cattle dataset.*** The cattle dataset was required from six freely roaming cows under normal living conditions, each cow's collar fitted with a triaxial accelerometer sampled at 25 Hz. In this study, the original acceleration time series was split using 2-second random sliding windows, resulting in a total of 10,429 data samples, each represented as a tensor of size $1 \times 3 \times 50$. Figure 8b illustrates the class distribution of these samples, which encompass five primary activities, i.e., grazing, moving, resting, ruminating, and salt-licking, accounting for 6%, 16%, 54%, 20%, and 4% of the dataset, respectively. The images depicting different behavioral states are shown in Fig. 9. It is evident that the dataset is imbalanced, with an imbalance ratio of 13.44.

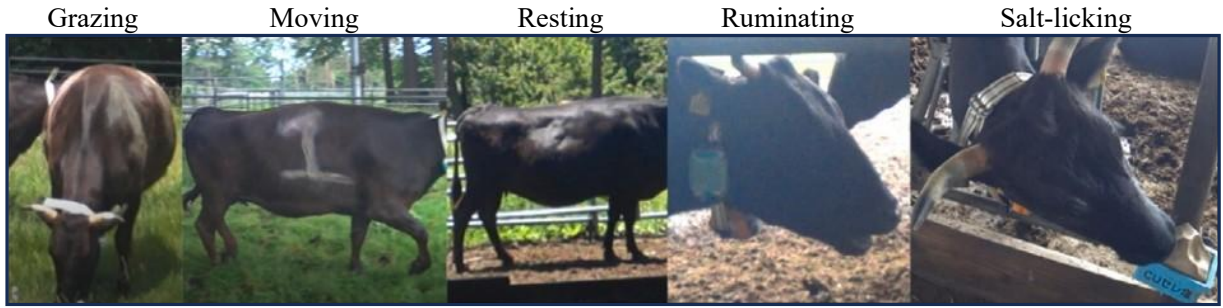

**Fig. 9**. Images of cattle with wearable sensors attached, capturing different behavioral states (Li et al., 2021).

***Horse dataset.*** The horse dataset comprises data collected from six horses equipped with neck-mounted triaxial accelerometers and triaxial gyroscopes, sampled at 100 Hz during both riding sessions and free pasture roaming over a seven-day period, as illustrated in Fig. 10. It contains a total of 87,621 labeled samples of 2-second duration, encompassing six distinct activities: eating, standing, trotting, galloping, walking-rider, and walking-natural. The sample distribution across these activities is imbalanced, with respective proportions of 18.32%, 5.84%, 28.62%, 4.50%, 38.94%, and 3.80%, corresponding to an imbalance ratio of 10.25 (Fig. 8c). Based on the findings from our previous study, which demonstrated that a sampling rate of 25 Hz yields optimal performance for horse activity recognition (Mao et al., 2023b), we downsampled all data to 25 Hz, resulting in a tensor structure of $1 \times 6 \times 50$ for each sample.

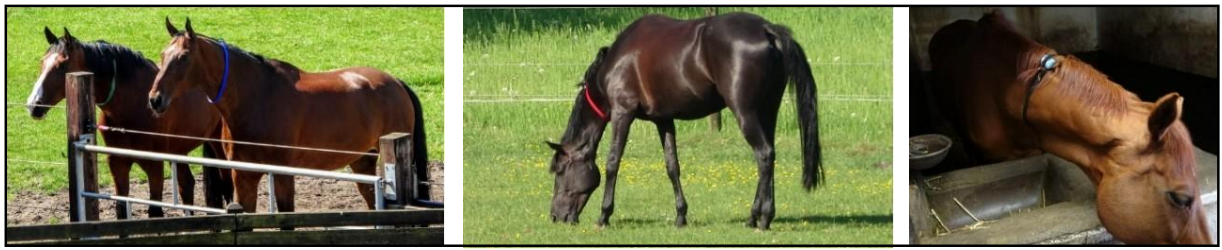

**Fig. 10**. Images of horse with wearable sensors attached (Kamminga et al., 2019).

### 3.2 Implementation details

The precision, recall, F1-score, and accuracy are utilized as metrics to assess the overall performance of the classification network. To validate the generalization capability of our proposed method, we utilize the leave-one-subject-out cross-validation technique (Mao et al., 2023b) on the goat dataset and horse dataset, where "one" refers to the data pertaining to a single individual. Considering that not all individual animals in the cattle dataset cover all behaviors, we adopt another commonly used validation method, i.e., stratified 5-fold cross-validation with non-overlapping splits (Hosseininoorbin et al., 2021), where samples of three, one, and one folds are separated as the training, validation, and testing datasets, respectively.

To ensure reproducibility and clarify the experimental setup, we provide a comprehensive summary of all relevant hyperparameters and their values in Table 1. Specifically, during training, an $L_2$ regularization term with weight decay values of $1 \times 10^{-4}$, $6 \times 10^{-2}$, and 0.1 for the goat, cattle, and horse datasets, respectively, is applied to the loss function to mitigate overfitting. We utilize the Adam optimizer with initial learning rates of $1 \times 10^{-4}$, $5 \times 10^{-4}$, and $1 \times 10^{-4}$ for the goat, cattle, and horse dataset respectively, and subsequently reduce the learning rates by a factor of 0.1 every 20 epochs. The training is configured with 100 epochs and a batch size of 256 across all datasets. The above values are determined based on configurations proven effective in our prior works (Mao et al., 2025, 2023b). We perform a grid search for hyperparameters $\tau$ in Eq. (2) and $k$ in Eq. (10) across the range of 0.1 to 1.0 with a step size of 0.1. The final values selected for the goat dataset are $\tau = 0.4$ and $k = 0.3$, for the cattle dataset are $\tau = 0.8$ and $k = 0.1$, and for the horse dataset are $\tau = 0.5$ and $k = 0.2$. The model exhibiting the highest validation accuracy is saved and subsequently evaluated using test data. For the goat and horse datasets, the baseline model are trained on data sampled at 12.5 Hz and 25 Hz, respectively, as these sampling rates have been established in prior studies to yield the highest performance for their respective recognition tasks (Mao et al., 2023b). For the cattle dataset, a sampling rate of 25 Hz is selected for training the baseline model, as it yields optimal performance in the preliminary experiments. To train the IBA-Net, we select three sampling rates for each dataset based on preliminary experiments indicating their complementary benefits for classifying different behaviors: 50 Hz, 25 Hz, and 12.5 Hz for the goat dataset; 25 Hz, 12.5 Hz, and 5 Hz for the cattle dataset; and 50 Hz, 25 Hz, and 12.5 Hz for the horse dataset. It is worth noting that our method is the first to integrate data from multiple sampling rates.

**Table 1.** Summary of hyperparameters employed in this study.

| Hyperparameters | Values | | |
| --- | --- | --- | --- |
| | Goat | Cattle | Horse |
| Weight decay | $1 \times 10^{-4}$ | $6 \times 10^{-2}$ | 0.1 |
| Learning rate | $1 \times 10^{-4}$ | $5 \times 10^{-4}$ | $1 \times 10^{-4}$ |
| Epoch | 100 | 100 | 100 |
| Batch size | 256 | 256 | 256 |
| $\tau$ | 0.4 | 0.8 | 0.5 |
| $k$ | 0.3 | 0.1 | 0.2 |
| Sampling rates (Hz) | 50, 25, 12.5 | 25, 12.5, 5 | 50, 25, 12.5 |

To validate our proposed method, we compare it against existing approaches that are trained solely on data with a single sampling rate and primarily focus on addressing the class imbalance problem. These approaches include resampling methods, i.e., KMeansSMOTE (KMS; Douzas et al., 2018) and RandomUnderSampler (RUS; Saripuddin et al., 2021), as well as reweighting methods, i.e., cost-sensitive cross-entropy loss (CS_CE; Khan et al., 2018), class-balanced focal loss (CB_FL; Cui et al., 2019), and Adaptive Class Suppression Loss (ACSL; Wang et al., 2021). All experiments are conducted using the PyTorch framework on a single NVIDIA GeForce RTX 3090 GPU. The source code for these experiments will be publicly available at https://github.com/Max-1234-hub/IBA-Net.

## 4. Results and discussion

Overall, the proposed IBA-Net outperforms existing approaches on the goat, cattle, and horse datasets. Ablation studies were conducted to validate the newly designed MFC and $NC^3$ modules, and the classification performance of these two modules for each behavior was thoroughly analyzed. Finally, the limitations of our study are discussed, and potential future work is proposed. The details are described as follows.

### 4.1. Performance comparisons

Tables 2, 3, and 4 demonstrate the comparison results of our methods on the goat, cattle, and horse datasets against several existing approaches, including resampling methods, i.e., KMS and RUS, and reweighting methods, i.e., CS_CE, CB_FL, and ACSL. The results show that our proposed IBA-Net achieves the best performance across all datasets, obtaining the highest metrics of 93.17% accuracy, 87.17% F1-score, and 91.71% recall for the goat dataset; 91.42% accuracy, 90.62% F1-score, 88.06% precision, and 93.64% recall for the cattle dataset; and 84.34% F1-score, 84.05% precision for the horse dataset. Particularly, the recall values show significant improvements, with increases of 16.98%, 15.78%, and 4.91% for the goat, cattle, and horse datasets respectively, compared to those of the baseline method. These improvements align with our initial expectation of achieving a high recall value and maximizing the classification accuracy across all behaviors. In addition, we observe that KMS obtains a limited improvement, which may be attributed to the fact that oversampling can lead to an over-representation of the minority class when increasing the number of minority samples, which in turn results in overfitting (Alkhawaldeh et al., 2023). It is worth noting that RUS leads to a sharp decline in classification performance. This is primarily because the dataset already has a limited number of samples, and further reducing the sample size through undersampling causes a significant loss of critical information (Saripuddin et al., 2021). Furthermore, the inclusion of CB_FL attains a higher improvement in recall values compared to other existing methods, which further confirms our choice of it for model optimization.

**Table 2.** Comparison of our methods with existing methods on the goat dataset.

| Method* | Accuracy (%) | F1-score (%) | Precision (%) | Recall (%) |
|---|---|---|---|---|
| Baseline | 92.13 | 77.52 | 86.82 | 74.73 |
| KMS (Douzas et al., 2018) | 92.29 | 81.10 | 79.85 | 82.92 |
| RUS (Saripuddin et al., 2021) | 81.00 | 71.48 | 74.93 | 70.16 |
| CS_CE (Khan et al., 2018) | 92.23 | 78.10 | **87.42** | 75.22 |
| CB_FL (Cui et al., 2019) | 91.38 | 81.88 | 78.18 | 87.35 |
| ACSL (Wang et al., 2021) | 92.51 | 77.21 | 87.38 | 73.79 |
| IBA-Net (Ours) | **93.17** | **87.17** | 83.80 | **91.71** |

* Baseline: The baseline refers to the model trained solely on data with a single sampling rate and optimized using the standard cross-entropy loss; KMS: KMeansSMOTE; RUS: RandomUnderSampler; CS_CE: Cost-sensitive Cross-Entropy loss; CB_FL: Class-balanced Focal Loss; ACSL: Adaptive Class Suppression Loss.

**Table 3.** Comparison of our methods with existing methods on the cattle dataset.

| Method* | Accuracy (%) | F1-score (%) | Precision (%) | Recall (%) |
|---|---|---|---|---|
| Baseline | 88.54 | 80.54 | 87.90 | 77.86 |
| KMS (Douzas et al., 2018) | 87.60 | 84.94 | 81.10 | 90.23 |
| RUS (Saripuddin et al., 2021) | 67.30 | 62.35 | 59.71 | 76.06 |
| CS_CE (Khan et al., 2018) | 88.55 | 80.53 | 87.90 | 77.81 |
| CB_FL (Cui et al., 2019) | 88.60 | 85.11 | 81.34 | 90.54 |
| ACSL (Wang et al., 2021) | 87.74 | 84.43 | 86.96 | 80.87 |
| IBA-Net (Ours) | **91.42** | **90.62** | **88.06** | **93.64** |

* Baseline: The baseline refers to the model trained solely on data with a single sampling rate and optimized using the standard cross-entropy loss; KMS: KMeansSMOTE; RUS: RandomUnderSampler; CS_CE: Cost-sensitive Cross-Entropy loss; CB_FL: Class-balanced Focal Loss; ACSL: Adaptive Class Suppression Loss.

**Table 4.** Comparison of our methods with existing methods on the horse dataset.

| Method* | Accuracy (%) | F1-score (%) | Precision (%) | Recall (%) |
|---|---|---|---|---|
| Baseline | 93.32 | 79.50 | 81.78 | 79.74 |
| KMS (Douzas et al., 2018) | 90.15 | 83.07 | 82.97 | 84.27 |
| RUS (Saripuddin et al., 2021) | 85.37 | 80.25 | 79.28 | **85.16** |
| CS_CE (Khan et al., 2018) | **93.42** | 79.54 | 81.72 | 80.03 |
| CB_FL (Cui et al., 2019) | 91.40 | 83.64 | 83.27 | 84.21 |
| ACSL (Wang et al., 2021) | 92.49 | 77.28 | 78.81 | 76.40 |
| IBA-Net (Ours) | 92.22 | **84.34** | **84.05** | 84.65 |

* Baseline: The baseline refers to the model trained solely on data with a single sampling rate and optimized using the standard cross-entropy loss; KMS: KMeansSMOTE; RUS: RandomUnderSampler; CS_CE: Cost-sensitive Cross-Entropy loss; CB_FL: Class-balanced Focal Loss; ACSL: Adaptive Class Suppression Loss.

### 4.2. Ablation studies

*4.2.1. Evaluation of the MFC and $NC^3$ modules*

The newly designed MFC and NC$^3$ modules serve as two essential components in our proposed IBA-Net to enhance classification performance. To demonstrate their contribution, we conduct experiments on the IBA-Net with and without these two modules, using the goat dataset, and present the results across various evaluation metrics in Table 4 along with the recall confusion matrices in Fig. 11. We can observe from Table 4 that the integration of the NC$^3$ module slightly increases the F1-score, precision, and recall by 0.51%, 0.29%, and 0.99%, respectively. Correspondingly, the classification accuracy of most behaviors has improved to varying degrees, including gains of 3.48% and 1.06% for the two minority classes of running and trotting, respectively, as evidenced by the comparison between Fig. 11b and 11a. This indicates that the NC$^3$ module is beneficial for enhancing the classification accuracy of minority classes. Notably, this also results in a slight misclassification of some grazing samples as running, which accounts for the minor decrease in overall classification accuracy as reflected in Table 5. In addition, the addition of the MFC module further magnifies the enhancement of the overall classification performance, with increments of 1.81%, 4.78%, 5.33%, and 3.37% in accuracy, F1-score, precision, and recall, respectively. Combined with the results presented in Fig. 11c, it can be observed that all activities possess desirable recall values, with most behaviors exhibiting an accuracy of over 90%. This is attributed to the MFC module's ability to capture the optimal movement patterns for each behavior by adaptively fusing features extracted from data at multiple sampling rates, thus guaranteeing high classification accuracy for each individual behavior.

**Table 5.** The performance of the proposed method with/without newly designed MFC and NC$^3$ modules.

| MFC | NC$^3$ | Accuracy (%) | F1-score (%) | Precision (%) | Recall (%) |
|---|---|---|---|---|---|
|  |  | 91.38 | 81.88 | 78.18 | 87.35 |
|  | √ | 91.33 | 82.39 | 78.47 | 88.34 |
| √ |  | **93.22** | 79.08 | 74.25 | 88.46 |
| √ | √ | 93.17 | **87.17** | **83.80** | **91.71** |

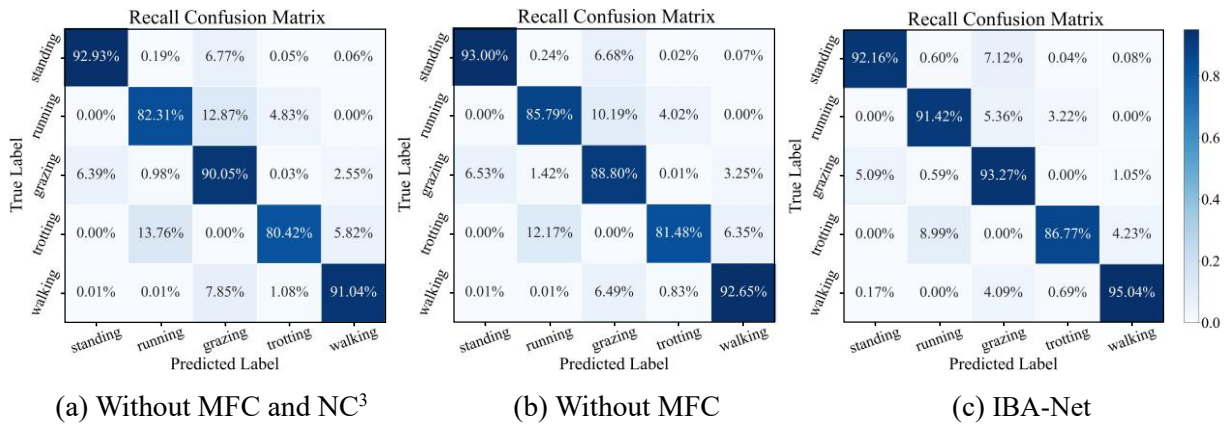

(a) Without MFC and NC$^3$      (b) Without MFC      (c) IBA-Net

**Fig. 11.** Confusion matrices for our IBA-Net (c), the variant without the MFC module (b), and the variant without both the MFC and NC$^3$ modules (a), respectively.

### 4.2.2. Analysis of the MFC module

***Benefits of soft weighted fusion for multiple features.*** A notable characteristic of the MFC module is that the integrated soft router can adaptively assign contribution degrees to features extracted at different sampling rates, with these degrees functioning as weights for soft feature fusion. To assess the effectiveness of this kind of soft-weighted fusion, we compare its performance with several normal fusion techniques, including addition, averaging, multiplication, and concatenation. The comparison results are presented in Table 6. Remarkably, our method of utilizing soft-weighted fusion outperforms all others in all evaluation metrics. This superior performance can be ascribed to the MFC module's ability to ensure that features obtained at the optimal sampling rate are maximized, thereby facilitating the capture of customized feature patterns for each individual behavior.

**Table 6.** Comparison of various fusion approaches for features extracted from multiple experts.

| Fusion method | Accuracy (%) | F1-score (%) | Precision (%) | Recall (%) |
|---|---|---|---|---|
| Addition | 91.87 | 80.20 | 75.75 | 90.02 |
| Averaging | 91.22 | 80.76 | 76.76 | 90.32 |
| Multiplication | 82.80 | 62.89 | 69.94 | 59.31 |
| Concatenation | 92.36 | 82.27 | 78.11 | 90.43 |
| Soft-weighted fusion (Ours) | **93.17** | **87.17** | **83.80** | **91.71** |

***Advantages of multiple feature fusion.*** Considering that it is difficult to ensure ideal performance for each individual behavior using a single sampling rate, the MFC module is proposed to fuse data from multiple sampling rates. To validate the MFC module, we compare its performance against methods that utilize a single sampling rate, including 50 Hz, 25 Hz, and 12.5 Hz. As shown in Table 7, our method with the MFC module obviously performs better than others across all evaluation metrics, confirming the advantages of multiple-feature fusion. We also visualize the two-dimensional embedded features from three expert networks before fusion, which are extracted from data sampled at three different rates, as well as the fused features in Fig. 12, utilizing t-distributed stochastic neighbor embedding (t-SNE) (Maaten and Hinton, 2008). When comparing Fig. 12d with Fig. 12a-c, it can be observed that the fused features exhibit more compact clusters within the same behavior categories and larger separations between different categories. This underscores the effectiveness of our proposed MFC module in deriving customized and discriminative features tailored to a specific behavior.

**Table 7.** Comparison of our IBA-Net trained on data from multiple sampling rates versus the model trained on data at a single sampling rate.

| Sampling rate | Accuracy (%) | F1-score (%) | Precision (%) | Recall (%) |
|---|---|---|---|---|
| 50Hz | 91.58 | 82.76 | 81.13 | 88.74 |
| 25Hz | 90.94 | 81.58 | 79.75 | 88.07 |
| 12.5Hz | 91.33 | 82.39 | 78.47 | 88.34 |
| Multiple sampling rates | **93.17** | **87.17** | **83.80** | **91.71** |

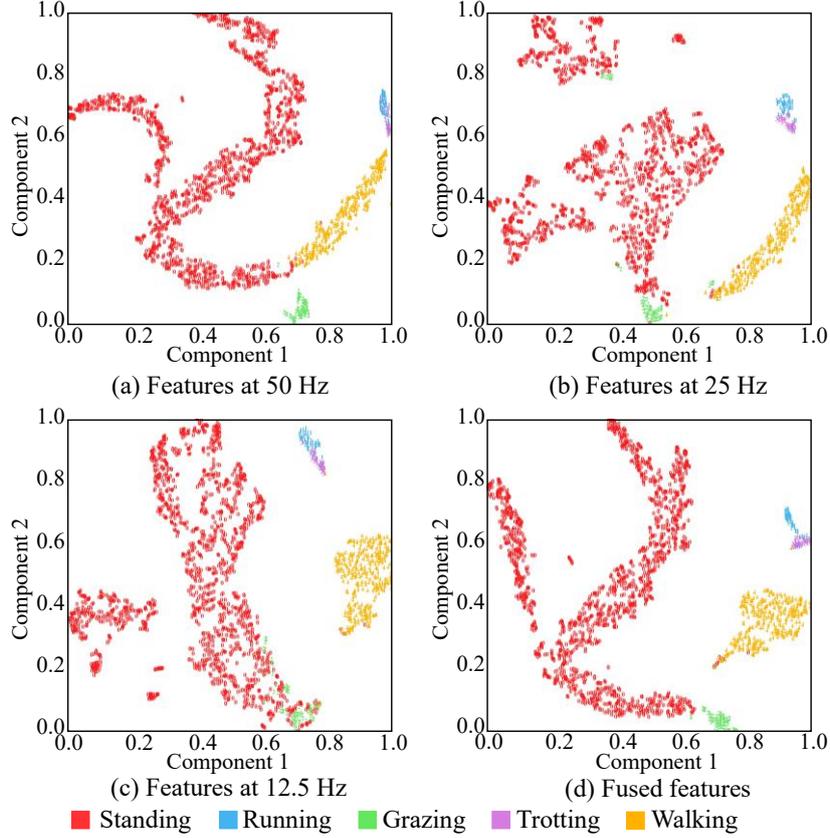

**Fig. 12**. t-SNE visualization of the feature vectors extracted from data at a single sampling rate data (a-c) and combined features derived from data at multiple sampling rates (d).

*4.2.3. Analysis of the NC³ module*

***Impact of the parameter k.*** The parameter $k$ in Eq. 10 controls the extent to which the ETF classification parameters are fixed in the final classification stage. A larger $k$ indicates a higher ratio of fixed classification parameters. We fix the parameter $\tau$ in Eq. 2 as 0.4 and compare the performance of our IBA-Net with various $k \in [0.1, 0.2, 0.3, 0.4, 0.5, 1.0]$, and the results are demonstrated in Table 8. The IBA-Net achieves its best performance when $k$ is set to 0.3, with accuracy, F1-score, precision, and recall of 93.17%, 87.17%, 83.80%, and 91.71%, respectively. As $k$ increases from 0.3, our model presents a gradual decline in performance. Notably, the IBA-Net exhibits extremely poor classification performance when $k$ is set to 1. This reveals that the complete fixation of classifier weights to their initial ETF state during the model training causes the network to lose its classification capability entirely.

**Table 8.** Performance comparison of our method with different values of $k$ in the MFC module.

| $k$ | Accuracy (%) | F1-score (%) | Precision (%) | Recall (%) |
|---|---|---|---|---|
| 0.1 | **93.17** | 84.25 | 81.10 | 88.94 |
| 0.2 | 93.11 | 86.09 | 82.19 | **91.73** |
| **0.3** | **93.17** | **87.17** | **83.80** | 91.71 |
| 0.4 | 91.96 | 83.39 | 79.20 | 90.12 |
| 0.5 | 92.20 | 70.36 | 69.22 | 77.79 |
| 1.0 | 21.96 | 22.81 | 32.23 | 27.41 |

*Visualization of the pair-wise angles between classifier vectors.* The magnitudes of the pair-wise angles between classifier weight vectors reflect the internal structure of the trained classifier in the embedding space. To validate the capability of the NC$^3$ module in maximizing these angles under class imbalance conditions, we visualize the pair-wise angles across five folds in Fig. 13, comparing our IBA-Net without and with the NC$^3$ module. It's evident that without the NC$^3$ module, the pair-wise angles exhibit substantial variation, with some measuring below 70° while others exceed 110°. This observation validates the assertion made by Fang et al. (2021) that classifier vectors of minority classes tend to shrink and converge toward each other in cases of class imbalance. In contrast, our IBA-Net with the NC$^3$ module significantly narrows the dispersion of pairwise angles, maintaining them consistently around 90°. This indicates that the NC$^3$ module effectively promotes the angles between classes to trend toward equality, aligning with the characteristics of ETF classifiers where angles are both maximized and equiangular (Papyan et al., 2020).

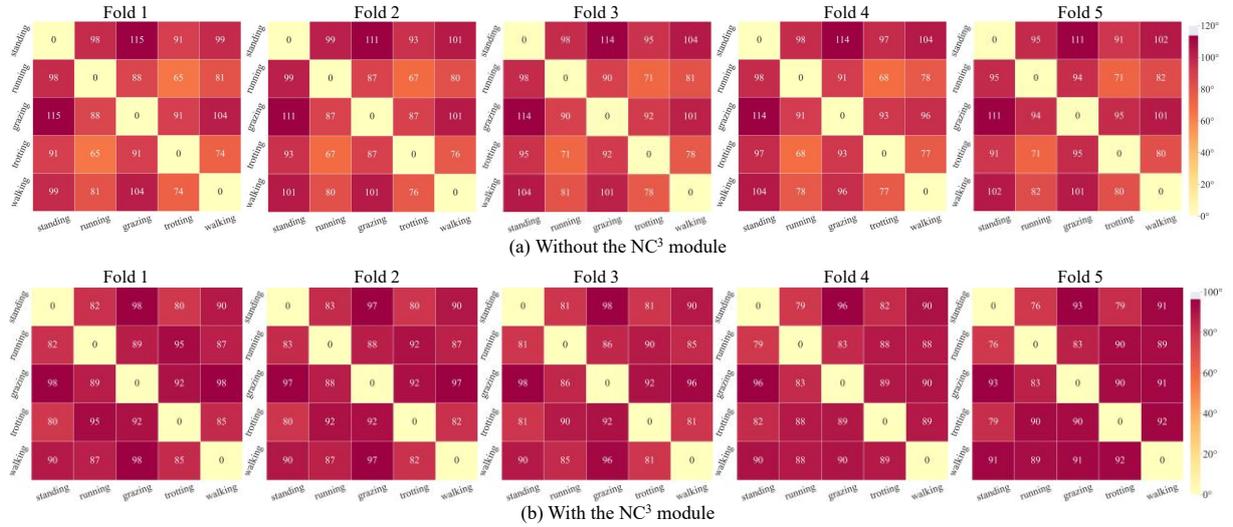

**Fig. 13**. Illustration of the pair-wise angles between the classifier weight vectors of our IBA-Net without (a) and with (b) the NC$^3$ module.

*Visualization of the training process.* To gain further insights into the impact of the NC$^3$ module during the training process, we randomly display the training and validation accuracy of our method, both without and with the NC$^3$ module, across two folds in Fig. 14. It is observable that our IBA-Net demonstrates superior training and validation accuracy, as well as a faster convergence rate. This further confirms the enhancement in model performance achieved by introducing fixed classifier weights adhering to the ETF classifier state.

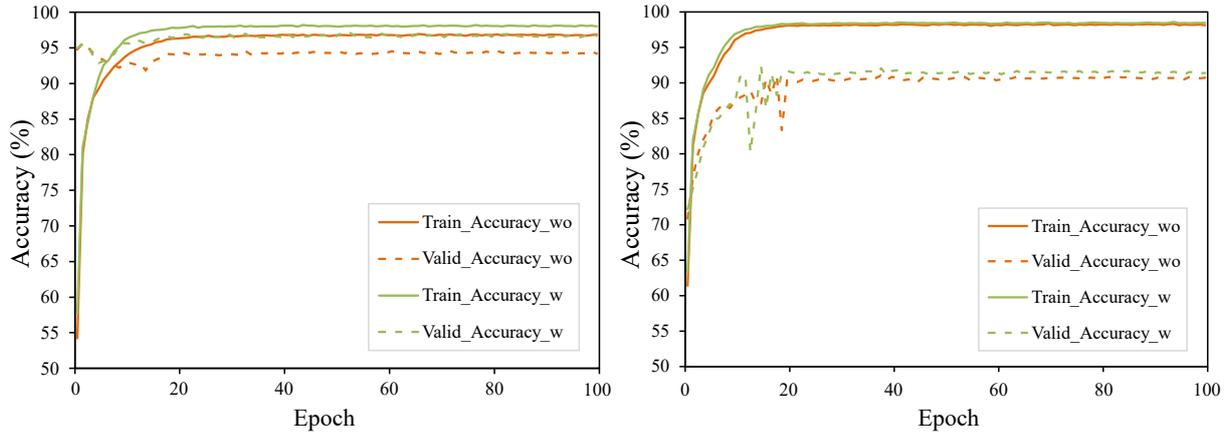

**Fig. 14.** The training and validation accuracy during the training process of our IBA-Net, both without and with the NC$^3$ module included.

### 4.2.4. Robustness against variations in class imbalance rate

The proposed IBA-Net introduces the ETF classifier as a regularization term into the original classification stage, enabling to maximization of the angles between pair-wise classifier weight vectors. To investigate the robustness of our method across different class imbalance rates, we elevate the imbalance of the goat dataset by downsampling the minority classes to 1/2 and 1/5 of their original numbers, thereby increasing the imbalance rate from its initial value of 98.05 to 2 and 5 times, respectively. The comparative classification performance of our method and the baseline is demonstrated in Table 9. We can observe that our IBA-Net consistently maintains superior performance across all evaluation metrics under varying degrees of class imbalance. Moreover, as the degree of imbalance increases, this advantage becomes more pronounced in terms of accuracy, F1-score, and precision. These notably demonstrate the robustness of our method in handling class imbalance scenarios.

**Table 9.** Performance comparison between our method and baseline under different degrees of imbalance conditions on the goat dataset.

| Imbalance rate | Method | Accuracy (%) | F1-score (%) | Precision (%) | Recall (%) |
|---|---|---|---|---|---|
| 98.05 | Baseline | 91.38 | 81.88 | 78.18 | 87.35 |
|  | IBA-Net | 93.17 | 87.17 | 83.80 | 91.71 |
| 196.1 | Baseline | 90.70 | 73.35 | 68.35 | 86.91 |
|  | IBA-Net | 92.60 | 77.60 | 72.78 | 89.26 |
| 490.25 | Baseline | 89.70 | 66.66 | 62.32 | 86.21 |
|  | IBA-Net | 92.23 | 75.27 | 69.82 | 88.52 |

### 4.3. Limitations and future work

The proposed approach integrates data from multiple sampling rates simultaneously and can adaptively extract the motion patterns from the optimal sampling rate data corresponding to a particular behavior. This significantly alleviates the dilemma of selecting the optimal sampling rate encountered in existing research. However, the necessity for multiple feature extraction networks to derive features from data sampled at varying rates inevitably leads to an increase in model size, elevating the computational cost for practical applications. Despite this, our method achieves an average inference time of less than 1.5 milliseconds per instance when tested on an Ultra7 3.80 GHz CPU using PyTorch with 16 threads and a batch size of 256, meeting the demand for real-time behavior prediction at normal sampling rates of up to 100 Hz. In addition, low-rank adaptation (LoRA) technology, which involves reducing the number of parameters by decomposing the original parameter matrix into the product of two low-rank matrices, has been proven effective in our previous research (Mao et al., 2025). Building upon this discovery, we intend to incorporate the LoRA architecture in our future work to replace our original feature extraction networks, with the aim of reducing the number of model parameters.

In terms of model architecture, the current study employs CNN-based encoders, a choice primarily grounded in their proven effectiveness in our prior work on activity recognition across multiple animal species, including cattle, sheep, and horses (Mao et al., 2025, 2023b, 2022b, 2021). While this approach has proven effective, we recognize that more specialized time-series modeling architectures may offer enhanced representational capacity. Notably, recent advances in temporal data processing have demonstrated the strong performance of models such as InceptionTime (Fawaz, 2020), ResNet-1D (Wang et al., 2017), and Transformers (Vaswani et al., 2017), which are specifically designed to capture multi-scale temporal patterns or long-range dependencies more effectively. Future work will explore integrating these advanced encoders to further enhance representational power and generalization ability.

While our proposed network demonstrates robust offline classification performance, its potential for near real-time animal behavior monitoring holds significant promise for advancing precision livestock farming. By integrating the proposed model with Internet of Things (IoT)-enabled wearable sensors, the system could enable timely detection of health-critical behaviors (e.g., lameness, abnormal feeding) and trigger immediate alerts for farmers. The MFC module's adaptive sampling rate selection aligns well with edge computing architectures, allowing computationally efficient feature extraction even under resource-constrained IoT devices. However, deployment in real-time scenarios requires further optimization of inference latency, particularly for high-frequency sensor data streams. Future work will explore lightweight model variants and edge-cloud collaborative frameworks to balance accuracy and responsiveness. Such an IoT-machine learning paradigm could revolutionize livestock management by merging real-time behavioral insights with automated decision-making, ultimately enhancing animal welfare and farm productivity.

On the other hand, the models developed in this study are tailored for a specific animal species, and we have not considered the generalization of these models to other species due to domain shifts. This is also a challenge confronted by the majority of current research, which directly constrains the applicability of the constructed models in real-world scenarios. Considering that large models like GPT (Radford et al., 2018) and SAM (Kirillov et al., 2023) can broadly learn from data of diverse origins and distributions, and exhibit exceptional performance, an intuitive and worthwhile direction to explore is developing a universal animal behavior recognition framework that is applicable to multiple species. This framework should be capable of flexibly capturing the unique behavioral patterns of different species, thereby enhancing the scalability and practicality of animal behavior recognition systems.

## 5. Conclusions

This study proposes a new approach named IBA-Net, which focuses on classification across all individual behaviors in farm animals by simultaneously customizing features and calibrating the classifier. On the one hand, an MFC module is designed to adaptively fuse data from different sampling rates and extract customized features tailored to various behaviors, effectively guaranteeing that each behavior obtains discriminative features. On the other hand, we devise an $NC^3$ module that integrates a fixed ETF classifier in the classification stage to maximize the angles between pair-wise classifier vectors, sufficiently ensuring that the classifier remains unbiased in classifying different behaviors as much as possible. The experimental results reflect that our proposed IBA-Net outperforms existing methods on all three wearable sensor datasets (goat, cattle, horse). Ablation studies are further carried out to analyze and confirm the effectiveness of the MFC and $NC^3$ modules, with a particular focus on their ability to enhance classification accuracy across all behaviors. In short, the promising performance of the proposed method demonstrates its potential to advance the field of wearable sensor-aided AAR.

**Acknowledgment**

This project was funded by the "National Natural Science Foundation of China" (Grant No. 62406094) and the "Research Startup Fund of Hangzhou Dianzi University" (Grant No. KYS085624257).

**Data availability statement**

The data used in this study are open-source datasets obtained from goats and cattle. The goat dataset can be accessed at: https://easy.dans.knaw.nl/ui/datasets/id/easy-dataset:78937/tab/2, while the cattle dataset is available at: https://zenodo.org/record/5849025#.ZE-y_3ZByHu.

## References

Alkhawaldeh, I.M., Albalkhi, I., Naswhan, A.J., 2023. Challenges and limitations of synthetic minority oversampling techniques in machine learning. World J. Methodol. 13, 373–378. https://doi.org/10.5662/wjm.v13.i5.373

Arablouei, R., Bishop-hurley, G.J., Bagnall, N., Ingham, A., 2024. Cattle behavior recognition from accelerometer data : Leveraging in-situ cross-device model learning. Comput. Electron. Agric. 227, 109546. https://doi.org/10.1016/j.compag.2024.109546

Arablouei, R., Currie, L., Kusy, B., Ingham, A., Greenwood, P.L., Bishop-Hurley, G., 2021. In-situ classification of cattle behavior using accelerometry data. Comput. Electron. Agric. 183, 106045. https://doi.org/10.1016/j.compag.2021.106045

Arablouei, R., Wang, L., Currie, L., Yates, J., Alvarenga, F.A.P., Bishop-Hurley, G.J., 2023. Animal behavior classification via deep learning on embedded systems. Comput. Electron. Agric. 207, 107707. https://doi.org/10.1016/j.compag.2023.107707

Bloch, V., Frondelius, L., Arcidiacono, C., Mancino, M., Pastell, M., 2023. Development and Analysis of a CNN- and Transfer-Learning-Based Classification Model for Automated Dairy Cow Feeding Behavior Recognition from Accelerometer Data. Sensors.

Cui, Y., Jia, M., Lin, T.Y., Song, Y., Belongie, S., 2019. Class-balanced loss based on effective number of samples, in: Proceedings of the IEEE Computer Society Conference on Computer Vision and Pattern Recognition. pp. 9268–9277. https://doi.org/10.1109/CVPR.2019.00949

Douzas, G., Bacao, F., Last, F., 2018. Improving imbalanced learning through a heuristic oversampling method based on k-means and SMOTE. Inf. Sci. (Ny). 465, 1–20. https://doi.org/10.1016/j.ins.2018.06.056

Eerdekens, A., Deruyck, M., Fontaine, J., Martens, L., De Poorter, E., Joseph, W., 2020. Automatic equine activity detection by convolutional neural networks using accelerometer data. Comput. Electron. Agric. 168, 105139. https://doi.org/10.1016/j.compag.2019.105139

Eerdekens, A., Deruyck, M., Fontaine, J., Martens, L., Poorter, E. De, Plets, D., Joseph, W., 2021. A framework for energy-efficient equine activity recognition with leg accelerometers. Comput. Electron. Agric. 183, 106020. https://doi.org/10.1016/j.compag.2021.106020

Fang, C., He, H., Long, Q., Su, W.J., 2021. Exploring deep neural networks via layer-peeled model: Minority collapse in imbalanced training. Proc. Natl. Acad. Sci. U. S. A. 118. https://doi.org/10.1073/pnas.2103091118

Fawaz, H.I., 2020. InceptionTime : Finding AlexNet for time series classification. Data Min. Knowl. Discov. 1936–1962. https://doi.org/10.1007/s10618-020-00710-y

Hendrycks, D., Gimpel, K., 2016. Gaussian Error Linear Units (GELUs) 1–10.

Hosseininoorbin, S., Layeghy, S., Kusy, B., Jurdak, R., Bishop-hurley, G., Portmann, M., 2021. Deep learning-based cattle activity classification using joint time-frequency data representation. Comput. Electron. Agric. 187, 106241. https://doi.org/10.1016/j.compag.2021.106241

Kamminga, J.W., Janßen, L.M., Meratnia, N., Havinga, P.J.M., 2019. Horsing around—A dataset comprising horse movement. Data 4, 1–13. https://doi.org/10.3390/data4040131

Kamminga, J.W., Le, D. V., Meijers, J.P., Bisby, H., Meratnia, N., Havinga, P.J.M., 2018. Robust sensor-orientation-independent feature selection for animal activity recognition on collar tags, in: Proceedings of the ACM on Interactive, Mobile, Wearable and Ubiquitous Technologies. pp. 1–27. https://doi.org/10.1145/3191747

Khan, S.H., Hayat, M., Bennamoun, M., Sohel, F.A., Togneri, R., 2018. Cost-sensitive learning of deep feature representations from imbalanced data. IEEE Trans. Neural Networks Learn. Syst. 29, 3573–3587. https://doi.org/10.1109/TNNLS.2017.2732482

Kim, J., Moon, N., 2022. Dog behavior recognition based on multimodal data from a camera and wearable device. Appl. Sci. 12, 3199. https://doi.org/10.3390/app12063199

Kirillov, A., Mintun, E., Ravi, N., Mao, H., Rolland, C., Gustafson, L., Xiao, T., Whitehead, S., Berg, A.C., Lo, W.-Y., Dollar, P., Girshick, R., 2023. Segment Anything, in: Proceedings of the IEEE International Conference on Computer Vision. pp. 4015–4026.

Kleanthous, N., Hussain, A., Khan, W., Sneddon, J., Liatsis, P., 2022. Deep transfer learning in sheep activity recognition using accelerometer data. Expert Syst. Appl. 207, 117925. https://doi.org/10.1016/j.eswa.2022.117925

Li, C., Tokgoz, K., Fukawa, M., Bartels, J., Ohashi, T., Takeda, K.I., Ito, H., 2021. Data augmentation for inertial sensor data in CNNs for cattle behavior classification. IEEE Sensors Lett. 5, 1–4. https://doi.org/10.1109/LSENS.2021.3119056

Li, Z., Shang, X., He, R., Lin, T., Wu, C., 2023. No Fear of Classifier Biases: Neural Collapse Inspired Federated Learning with Synthetic and Fixed Classifier.

Liseune, A., den Poel, D. Van, Hut, P.R., van Eerdenburg, F.J.C.M., Hostens, M., 2021. Leveraging sequential information from multivariate behavioral sensor data to predict the moment of calving in dairy cattle using deep learning. Comput. Electron. Agric. 191, 106566. https://doi.org/10.1016/j.compag.2021.106566

Maaten, L. van der, Hinton, G., 2008. Visualizing data using t-sne. J. Mach. Learn. Res. 9, 2579–2605. https://doi.org/10.1007/s10479-011-0841-3

Mao, A., Giraudet, C., Liu, K., De, I., Nolasco, A., Xie, Zhiqin, Xie, Zhixun, Gao, Y., Theobald, J., Bhatta, D., Stewart, R., Mcelligott, A.G., 2022a. Automated identification of chicken distress vocalizations using deep learning models. J. R. Soc. Interface 19, 20210921.

Mao, A., Huang, E., Gan, H., 2022b. FedAAR : A novel federated learning framework for animal activity recognition with wearable sensors. Animals 12, 2142.

Mao, A., Huang, E., Gan, H., Parkes, R.S. V, Xu, W., 2021. Cross-modality interaction network for equine activity recognition using imbalanced multi-modal data †. Sensors 21, 5818.

Mao, A., Huang, E., Wang, X., Liu, K., 2023a. Deep learning-based animal activity recognition with wearable sensors: Overview, challenges, and future directions. Comput. Electron. Agric. 211, 108043. https://doi.org/10.1016/j.compag.2023.108043

Mao, A., Zhu, M., Guo, Z., He, Z., Norton, T., Liu, K., 2025. CKSP: Cross-species Knowledge Sharing and Preserving for Universal Animal Activity Recognition. Biosyst. Eng. 259, 104303. https://doi.org/10.1016/j.biosystemseng.2025.104303

Mao, A., Zhu, M., Huang, E., Yao, X., Liu, K., 2023b. A teacher-to-student information recovery method toward energy-efficient animal activity recognition at low sampling rates. Comput. Electron. Agric. 213, 108242. https://doi.org/10.1016/j.compag.2023.108242

Pan, Z., Chen, H., Zhong, W., Wang, A., Zheng, C., 2023. A CNN-based Animal Behavior Recognition Algorithm for Wearable Devices. IEEE Sens. J. 23, 1–1. https://doi.org/10.1109/jsen.2023.3239015

Papyan, V., Han, X.Y., Donoho, D.L., 2020. Prevalence of Neural Collapse during the terminal phase of deep learning training arXiv:1707, 1–17. https://doi.org/10.1073/pnas.XXXXXXXXXX

Radford, A., Narasimhan, K., Salimans, T., Sutskever, I., 2018. Improving Language Understanding by Generative Pre-Training. https://doi.org/10.4310/HHA.2007.v9.n1.a16

Saripuddin, M., Suliman, A., Syarmila Sameon, S., Jorgensen, B.N., 2021. Random Undersampling on Imbalance Time Series Data for Anomaly Detection. ACM Int. Conf. Proceeding Ser. 151–156. https://doi.org/10.1145/3490725.3490748

Suh, S., Lee, H., Lukowicz, P., Oh, Y., 2021. CEGAN : Classification enhancement generative adversarial networks for unraveling data imbalance problems. Neural Networks 133, 69–86. https://doi.org/10.1016/j.neunet.2020.10.004

Vaswani, A., Shazeer, N., Parmar, N., Uszkoreit, J., Jones, L., Gomez, A.N., Kaiser, Ł., Polosukhin, I., 2017. Attention is all you need. Adv. Neural Inf. Process. Syst. 2017-Decem, 5999–6009.

Vural, V., Dy, J.G., 2004. A Hierarchical Method for Multi-Class Support Vector Machines, in: Proceedings of the Twenty-First International Conference on Machine Learning. p. 105. https://doi.org/10.1145/1015330.1015427

Walton, E., Casey, C., Mitsch, J., Vázquez-Diosdado, J.A., Yan, J., Dottorini, T., Ellis, K.A., Winterlich, A., Kaler, J., 2018. Evaluation of sampling frequency, window size and sensor position for classification of sheep behaviour. R. Soc. Open Sci. 5, 171442. https://doi.org/10.1098/rsos.171442

Wang, T., Zhu, Y., Zhao, C., Zeng, W., Wang, J., Tang, M., 2021. Adaptive class suppression loss for long-tail object detection, in: Proceedings of the IEEE/CVF Conference on Computer Vision and Pattern Recognition. pp. 3103–3112.

Wang, Z., Yan, W., Oates, T., 2017. Time Series Classification from Scratch with Deep Neural Networks : A Strong Baseline, in: International Joint Conference on Neural Networks.

Zarka, J., Guth, F., Mallat, S., 2021. Separation and Concentration in Deep Networks. ICLR 2021 - 9th Int. Conf. Learn. Represent. 1–16.

Zhang, S., Li, Z., Yan, S., He, X., Sun, J., 2021. Distribution alignment: A unified framework for long-tail visual recognition, in: Proceedings of the IEEE/CVF Conference on Computer Vision and Pattern Recognition. pp. 2361–2370.